\documentclass[a4paper]{journal}
\usepackage[a4paper, margin=1in]{geometry}
\usepackage{multicol}
\usepackage{parskip}
\usepackage{titling}
\usepackage{indentfirst}
\usepackage{authblk}
\title{\textbf{{\LARGE  Technical Opinion: From Animal Behaviour to Autonomous Robots}}}
\date{}

\author[1,2]{Chinedu Pascal Ezenkwu}

\author[1]{Andrew Starkey}
\affil[1]{School of Engineering, University of Aberdeen, United Kingdom}
\affil[2]{Electrical/Electronic \& Computer Engineering Department, University of Uyo, Nigeria}

\begin{document}
\maketitle
\begin{abstract} 

  With the rising applications of robots in unstructured real-world environments, roboticists are increasingly concerned with the problems posed by the complexity of such environments. One solution to these problems is robot autonomy. Since nature has already solved the problem of autonomy it can be a suitable model for developing autonomous robots. This paper presents a concise review on robot autonomy from the perspective of animal behaviour. It examines some state-of-the-art techniques as well as suggesting possible research directions. 

\end{abstract}

\section*{Introduction}


As the scope for robotic applications extends from structured to unstructured and more complex environments, autonomy has become a desideratum for most of today's robots. The practice of handcrafting robots does not give them the capability to cope with unforeseen situations. Although several research contributions have been made towards robot autonomy, we are nowhere near the level of autonomy that is exhibited by animals, even ones at the lowest biological level of organisation. This is because animals are born with innate capabilities, both in their body structure and intelligence, to survive and develop in their milieus; their behaviours and sometimes their morphological traits can evolve to adapt to persistent changes in their habitats. For example, Corcoran \textit{et al} \cite{corcoran2011tiger} studied the co-evolutionary battle between the bat and the moth. Bats rely on echolocation for finding their prey. They produce high frequency sounds and examine the echoes for estimating the positions of their prey. Since moths have been on the bat's favourite menu for tens of millions of years, they have evolved simple ears to detect possible attacks, produce their own ultrasounds to jam the sound from the predator and initiate amazing deflective strategies for manoeuvres. This ability of the moth to develop skills that were not originally there in order to survive in its environment is a fundamental product of autonomy. We posit that this idea of autonomy should remain the same in robotics. An autonomous robot, without any need for external change of the underlying algorithms, should learn to develop novel skills in order to cope with unpredictable situations.

In our previous review, we defined what it means for an agent to be understood as autonomous in terms of specified attributes \cite{ezenkwu2019machine}. The attributes are classified into two categories: low-level attributes and high-level attributes. The low-level attributes constitute the separating line between autonomous and automatic or automated agents. They include perception, actuation, learning, context-awareness and decision-making. We argue that any system, whether biological or artificial, has to possess at least these low-level attributes to be called autonomous. The high-level attributes of autonomy are the advanced attributes of autonomy. They include domain-independence, self-motivation, self-recovery and self-identification
of goals. While the low-level attributes have been extensively addressed in robotics and artificial intelligence, the high-level attributes remain a challenge. In the following section, we discuss why some popular AI techniques such as deep learning and reinforcement learning are not suitable methods for achieving high-level autonomy. Furthermore, some techniques that agree with our definition of autonomous behaviour in robots are highlighted.   

\section*{Review of State-of-the-art Techniques}
 Inspired by animal behaviours or nature in general, AI is developing at a mind-boggling speed. Some of the more sophisticated AI techniques have shown promising results in different areas of applications. For example, in 2015 DeepMind Technologies developed AlphaGo, a deep reinforcement learning algorithm that became the first to defeat a master in the game of Go; deep learning has also proven to be the go-to method in different classes of AI problems such as computer vision, natural language processing (NLP), self-driving cars  and so on. Despite the level of progress attained by these methods in various application areas we believe that these methods on their own are not suitable for high degree autonomy.
  
 According to Goodfellow \cite{goodfellow2016deep} ``some artificial neural networks have nearly as many connections per neuron as a cat, and it is quite common for other neural networks to have as many connections per neuron as smaller mammals like mice''. Revisiting the moth example: although moths have fewer connections per neuron than some artificial neural networks, they posses a higher degree of autonomy than the latter. A more radical example is the behaviour of the trichoplax, a small disk-shaped marine metazoan which exhibits a sequence of coordinated feeding and response behaviour despite having no nervous system \cite{smith2015coordinated}. This arouses curiosity with regard to the nature of autonomy in animals. While many improvements have been made in AI in recent times by creating neural networks with massive interconnections of neurons in mimicry of the human brain, it can be argued that these massive interconnections alone are not enough to realise common sense and autonomous behaviours in robots. Although different classes of animals have different neural complexities, they exhibit a high degree of autonomy in their milieus. So, instead of focusing only on the anatomical similarities between the AI and biological agents, it is also good to consider the basic theories and attributes that underpin autonomous behaviours in animals and seek to formalise those in artificial systems. 

The emergence of deep learning has revolutionalised how AI is applied in virtually all application domains especially in perceptual tasks such as visual recognition, speech recognition and so on. However, the concept of deep learning does not agree with our ideas of autonomous behaviour. Most successful deep learning algorithms like convolutional neural networks (CNN) and recurrent neural networks (RNN) are supervised learning algorithms, and therefore they require a teacher to provide them with accurate labels of sensory inputs. This makes them impracticable in unknown or totally new environments. An autonomous agent should be driven by self-motivation to learn and improve in unknown environments without a teacher. Moreover, deep learning algorithms are data and resource hungry. For example, with 1.2 million labelled training examples, trained on two GTX 580 3GB GPUs, AlexNet, a deep convolutional neural network that won the 2010 ImageNet Large-Scale Visual Recognition Challenge (ILSVRC), took between five and six days to learn how to classify images into 1000 different categories \cite{krizhevsky2012imagenet}. Humans, and animals like rats and primates, do not require so many examples to learn \cite{aggleton1985one} and as a result they can quickly adapt to changes in their environments without having to wait for too many experiences.

One of the most applied AI techniques in autonomous decision-making is reinforcement learning (RL). Amongst all machine learning paradigms, RL has the most obvious similarity with goal-directed behaviour in animals. The modern RL can trace its roots back to the psychology of animal learning, specifically Thorndike's classic theory of the law of effect which suggests that animals tend to increase behaviours that are followed by pleasant consequences and decrease behaviours that are followed by negative consequences \cite{thorndike1911animal}. RL has been successful in a number of problems from computer games to self-flying helicopters. However, a major challenge of applying RL in autonomous robotic scenarios is that it requires handcrafting of the reward function, which is often difficult to design for some application scenarios. As such, RL depends on demonstration data from human experts for most real-life applications; this data is non-existent in most cases. Moreover, reward functions are often environment-dependent, causing the algorithms to find it difficult to cope when environments change in ways that are not captured in the reward functions. RL exists in two categories i.e. model-free and model-based RL. This classification is dependent on whether the transition model of the environment is required or not. Unlike model-based RL, model-free RL does not require the model of the environment to learn. Although model-based RL has proven to be sample efficient and has high generalisation, it is not often applied in most practical problems because environment models are difficult to design for unpredictable or unknown environments. Also, model-based RL is strongly coupled with environment models and they perform worse than model-free methods when these environment models no longer capture the current situation, unless the designer manually updates them. As a result of this, model-free RL is the most applied RL method in real-world contexts. However, these methods are sample inefficient and have a very poor generalisation performance; they take a long time to adapt to changing goals and/or environments.
\section*{The Future of Autonomous Learning}

A number of research areas such as developmental robotics \cite{cangelosi2015developmental}, enactive learning \cite{georgeon2013enactive} and unsupervised learning \cite{ezenkwu2019unsupervised} seem to offer good alternatives for autonomous robots. Developmental robotics is inspired by the cognitive development in human infants. This is in line with Alan Turing's idea that it is easier to build an artificial baby and train it to maturity than trying to build and simulate an adult mind \cite{turing2009computing}. Enactive learning seeks to develop a self-motivated agent that has the capability to self-program itself while engaging in active perception. In our ongoing research, we seek to represent causation in unsupervised self-organising neural networks. This will help an agent to learn the sensorimotor map of its environment for planning instead of depending on the designer to formulate it as in model-based RL. While we believe that these methods have the potential to drive autonomous robotic research, we also agree that no single method has the capabilities to realise a truly autonomous behaviour. However, a combination of two or more methods will yield better results to the challenges of designing autonomous robots.
\section*{Conclusions}

This paper has reviewed robot autonomy in the context of animal behaviour. It argues that the popular state-of-the-art AI methods, on their own, are not capable of achieving high degree autonomy. Some AI methods that seem to agree with our definition of autonomy are briefly explained. This paper does not claim that any single method has the capability to achieve high degree autonomy but understanding the strengths and weaknesses of different methods can help in synergising them for autonomous robot applications.


\begin{thebibliography}{100} 
	\bibitem{corcoran2011tiger} Aaron J Corcoran, Jesse R Barber, Nickolay I Hristov, and William E Conner. How do
	tiger moths jam bat sonar? Journal of Experimental Biology, 214(14):2416{2425, 2011}.
	
	\bibitem{ezenkwu2019machine} Chinedu Pascal Ezenkwu and Andrew Starkey. Machine autonomy: Definition,
		approaches, challenges and research gaps. In Intelligent Computing-Proceedings of
		the Computing Conference, pages 335–358. Springer, 2019.
	\bibitem{goodfellow2016deep} Ian Goodfellow, Yoshua Bengio, and Aaron Courville. Deep learning. MIT press,
	2016.
	
	\bibitem{smith2015coordinated}Carolyn L Smith, Natalia Pivovarova, and Thomas S Reese. Coordinated feeding
	behavior in trichoplax, an animal without synapses. PloS one, 10(9), 2015.
	
	\bibitem{krizhevsky2012imagenet} Alex Krizhevsky, Ilya Sutskever, and Geoffrey E Hinton. Imagenet classification
	with deep convolutional neural networks. In Advances in neural information processing systems, pages 1097–1105, 2012.
	
	\bibitem{aggleton1985one} JP Aggleton. One-trial object recognition by rats. The Quarterly Journal of
	Experimental Psychology, 37(4):279–294, 1985.
	
	\bibitem{thorndike1911animal} EL Thorndike. Animal intelligence; experimental studies. on cover: the animal
	behavior series, 1911.
	
	\bibitem{cangelosi2015developmental}  Angelo Cangelosi and Matthew Schlesinger. Developmental robotics: From babies
	to robots. MIT press, 2015.
	
	\bibitem{georgeon2013enactive} Olivier L Georgeon, Christian Wolf, and Simon Gay. An enactive approach to
	autonomous agent and robot learning. In 2013 IEEE Third Joint International
	Conference on Development and Learning and Epigenetic Robotics (ICDL), pages
	1–6. IEEE, 2013.
	
	\bibitem{ezenkwu2019unsupervised}  Chinedu Pascal Ezenkwu and Andrew Starkey. Unsupervised temporospatial neural architecture for sensorimotor map learning. IEEE Transactions on Cognitive
	and Developmental Systems, 2019.

	
	\bibitem{turing2009computing} Alan M Turing. Computing machinery and intelligence. In Parsing the Turing
	Test, pages 23–65. Springer, 2009.
	
	
\end{thebibliography}

\end{document}